\documentclass[10pt,twocolumn,letterpaper]{article}

\usepackage[accsupp]{axessibility}  
\usepackage{iccv}
\usepackage{times}
\usepackage{epsfig}
\usepackage{graphicx}
\usepackage{amsmath}
\usepackage{amssymb}

\usepackage{booktabs}
\usepackage{multirow}
\usepackage{caption}
\usepackage{everyshi}
\usepackage[pagebackref=true,breaklinks=true,letterpaper=true,colorlinks,bookmarks=false]{hyperref}

\iccvfinalcopy 

\def\iccvPaperID{5411} 

\ificcvfinal\fi

\def\ie{\emph{i.e.,~}}

\def\etal{{\em et al.}}
\newcommand{\supp}{\emph{supplementary material}}

\ifdefined \GramaCheck
  \newcommand{\CheckRmv}[1]{}
  \newcommand{\figref}[1]{Figure 1}%
  \newcommand{\tabref}[1]{Table 1}%
  \newcommand{\secref}[1]{Section 1}
  \newcommand{\algref}[1]{Algorithm 1}
  \renewcommand{\eqref}[1]{Equation 1}
\else
  \newcommand{\CheckRmv}[1]{#1}
  \newcommand{\figref}[1]{Figure~\ref{#1}}
  \newcommand{\tabref}[1]{Table~\ref{#1}}
  \newcommand{\secref}[1]{Section~\ref{#1}}
  \newcommand{\algref}[1]{Alg.~\ref{#1}}
  \renewcommand{\eqref}[1]{(\ref{#1})}
\fi

\newcommand{\ourM}{{MonoNeRF}}
\newcommand{\DynNeRF}{{DynNeRF}}

\begin{document}

\title{\ourM: Learning a Generalizable Dynamic Radiance Field \\
from Monocular Videos}

\author{Fengrui Tian$^{1}$ \quad Shaoyi Du$^{1}$ \quad Yueqi Duan$^{2\dagger}$\\
$^{1}$National Key Laboratory of Human-Machine Hybrid Augmented Intelligence, \\
National Engineering Research Center for Visual Information and Applications, \\ 
and Institute of Artificial Intelligence and Robotics, Xi'an Jiaotong University \\
$^{2}$Department of Electronic Engineering, Tsinghua University \\
{\tt\small tianfr@stu.xjtu.edu.cn, dushaoyi@gmail.com, duanyueqi@tsinghua.edu.cn}
}

\twocolumn[{%
\maketitle
    \begin{center}
		\centering
        \CheckRmv{
    \includegraphics[width=.95\linewidth]{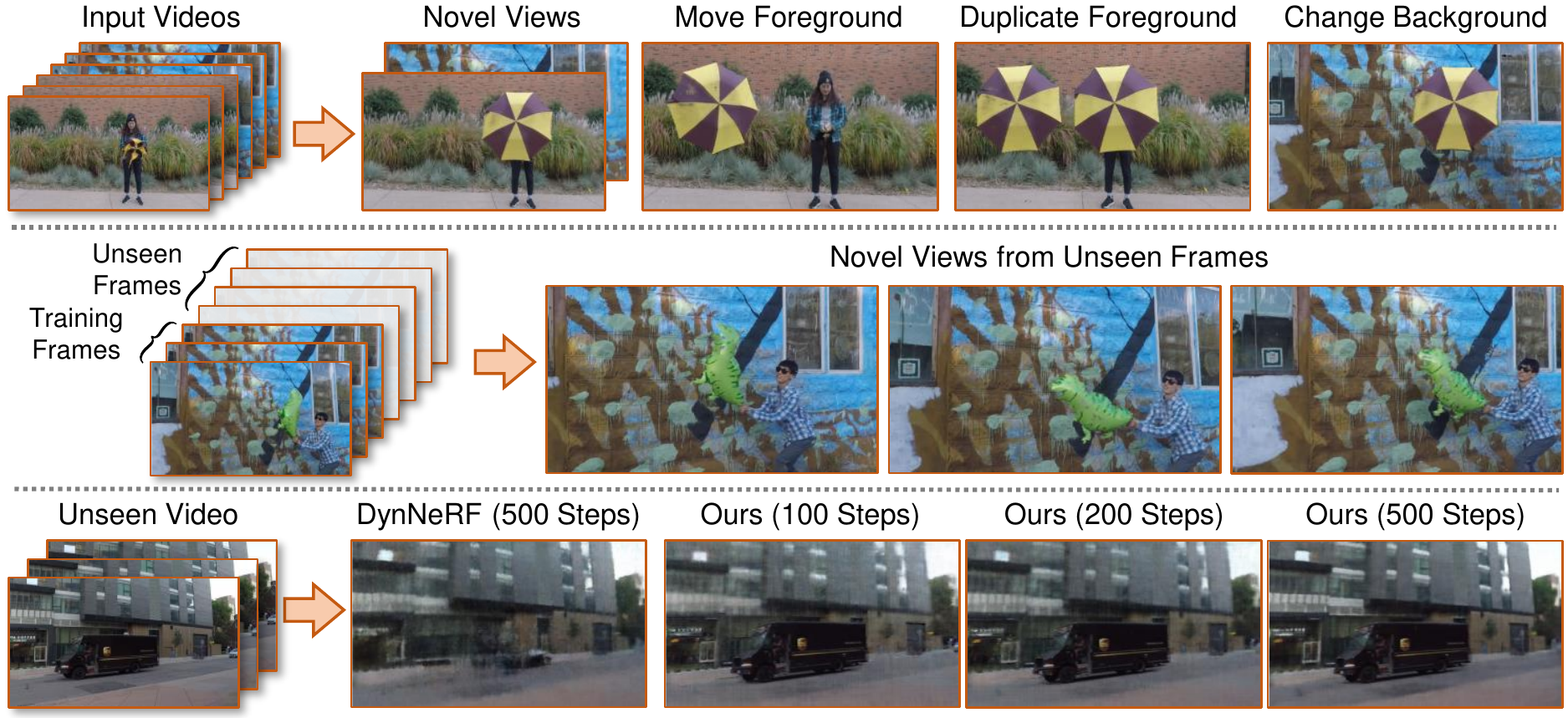}
\captionof{figure}{
\ourM~learns a generalizable dynamic radiance field from multiple monocular videos. 
While video-based NeRF learns the field from positional encoding,
we propose to learn from the extracted video features, which are generalizable to scenes. In this way, \ourM~supports scene editing applications (\textbf{Top}) and unseen frame synthesis (\textbf{Middle}). Compared with \DynNeRF~\cite{21iccv/chen_dynerf}, \ourM~could adapt to novel scenes (\textbf{Bottom}) with hundreds of fine-tuning steps (about 10 minutes). 
	}
}
\label{fig:teaser}

\end{center}	
\label{fig:teaser_main}
}]
\renewcommand{\thefootnote}{}
\footnotetext{$^\dagger$: Corresponding author.}
\begin{abstract}
    In this paper, we target at the problem of learning a generalizable dynamic radiance field from monocular videos. Different from most existing NeRF methods that are based on multiple views, monocular videos only contain one view at each timestamp, thereby suffering from ambiguity along the view direction in estimating point features and scene flows. Previous studies such as \DynNeRF~disambiguate point features by positional encoding, which is not transferable and severely limits the generalization ability. As a result, these methods have to train one independent model for each scene and suffer from heavy computational costs when applying to increasing monocular videos in real-world applications. To address this, We propose \ourM~to simultaneously learn point features and scene flows with point trajectory and feature correspondence constraints across frames. More specifically, we learn an implicit velocity field to estimate point trajectory from temporal features with Neural ODE, which is followed by a flow-based feature aggregation module to obtain spatial features along the point trajectory. We jointly optimize temporal and spatial features in an end-to-end manner. Experiments show that our \ourM~is able to learn from multiple scenes and support new applications such as scene editing, unseen frame synthesis, and fast novel scene adaptation. Codes are available at \url{https://github.com/tianfr/MonoNeRF}.
\end{abstract}

\section{Introduction}
\label{sec:intro}

Novel view synthesis \cite{93cgit/chen_nvs} is a highly challenging problem. It facilitates many important applications in movie production, sports event, and virtual reality.
The long standing problem recently has witnessed impressive progress due to the neural rendering
technology \cite{20eccv/ben_nerf,21cvpr/ricardo_nerfw,22cvpr/kangle_dsnerf}.
Neural Radiance Field (NeRF) \cite{20eccv/ben_nerf,21cvpr/yu_pixelnerf,20nips/liu_neuralsparse,22cvpr/nerfindark,21cvpr/neus,22eccv/xiangli_bungeenerf,22cvpr/xu_pointnerf}
shows
that photo-realistic scenes can be represented by an implicit neural network.
Concretely, taken as a query the position and viewing direction of the posed image, the network outputs the color of each pixel by volume rendering method.
Among these approaches, it is supposed that the scene is static and 
can be observed from 
multiple views at the same time. Such assumptions are violated by numerous videos
uploaded to the Internet, which usually contain dynamic foregrounds, recorded
by the monocular camera.

More recently, some studies aim to explore how to learn dynamic radiance field from monocular videos \cite{21iccv/du_nerflow,21iccv/chen_dynerf,21cvpr/li_nsff,22nips/wu_d2nerf,21cvpr/xian_stnif,21iccv/tretschk_nr-nerf}. Novel view synthesis from monocular videos is a challenging task. As foreground usually dynamically changes in a video, there is ambiguity in the view direction to estimate precise point features and dense object motions (\ie~scene flow \cite{99iccv/vedula_sceneflow}) from single views. In other words, we can only extract the projected 2D intra-frame local features and inter-frame optical flows, but fail to obtain precise 3D estimations.
Previous works address this challenge by representing points as 4D position features (coordinate and time), so that the learned positional encoding provides specific information for each 3D point in the space \cite{21iccv/du_nerflow,21iccv/chen_dynerf,21cvpr/li_nsff,22nips/wu_d2nerf,21cvpr/xian_stnif,21iccv/tretschk_nr-nerf}. Based on positional encoding, these methods make efforts on exploiting scene priors~\cite{21cvpr/guy_nerface,22cvpr/weng_humannerf,22nips/wu_d2nerf} or adding spatio-temporal regularization~\cite{21iccv/du_nerflow,21iccv/chen_dynerf,21cvpr/li_nsff,21cvpr/xian_stnif} to learn a more accurate dynamic radiance field.


However, while positional encoding successfully disambiguates 3D points from monocular 2D projections, it severely overfits to the training video clip and is not transferable. Therefore, existing positional encoding based methods have to optimize one independent model for each dynamic scene. With the fast increase of monocular videos in reality, they suffer from heavy computational costs and lengthy training time to learn from multiple dynamic scenes. Also, the lack of generalization ability limits further applications of scene editing which requires interaction among different scenes. 
A natural question is raised: can we learn a generalizable dynamic radiance field from monocular videos?


In this paper, we provide a positive answer to this question. The key challenge of this task is to learn to extract generalizable point features in the 3D space from monocular videos. While independently using 2D local features and optical flows suffers from ambiguity along the ray direction, they provide complementary constraints to jointly learn 3D point features and scene flows. On the one hand, for the sampled points on each ray, optical flow provides generalizable constraints that limit the relations of their point trajectories. On the other hand, for the flowing points on each estimated point trajectory, we consider that they should share the same point features. We estimate each point feature by aggregating their projected 2D local features, and design feature correspondence constraints to correct unreasonable trajectories.


To achieve this, we propose \ourM~to build a generalizable dynamic radiance field for multiple dynamic scenes. 
We hypothesize that a point moving along its trajectory over time keeps the consistent point feature. Our method concurrently predicts 3D point features and scene flows
with point trajectory and feature correspondence constraints in monocular video frames. More specifically, we first propose to learn an implicit velocity field that encodes the speed from the temporal feature of each point. We supervise the velocity field with optical flow and integrate continuous point trajectories on the field with Neural ODE \cite{18nips/chen_neuralode}. Then, we propose a flow-based feature aggregation module to sample spatial features of each point along the point trajectory. We incorporate the spatial and temporal features as the point feature to query the color and density for image rendering and jointly optimize point features and trajectories in an end-to-end manner. As shown in \figref{fig:teaser}, experiments demonstrate that our \ourM~is able to render novel views from multiple dynamic videos and support new applications such as scene editing, unseen frame synthesis, and fast novel scene adaption. Also, in the widely-used setting of novel view synthesis on training frames from single videos, our \ourM~still achieves better performance than existing methods despite that cross-scene generalization ability is not required in this setting.

\CheckRmv{
\begin{figure*}[t]
  \centering
   \includegraphics[width=\linewidth]{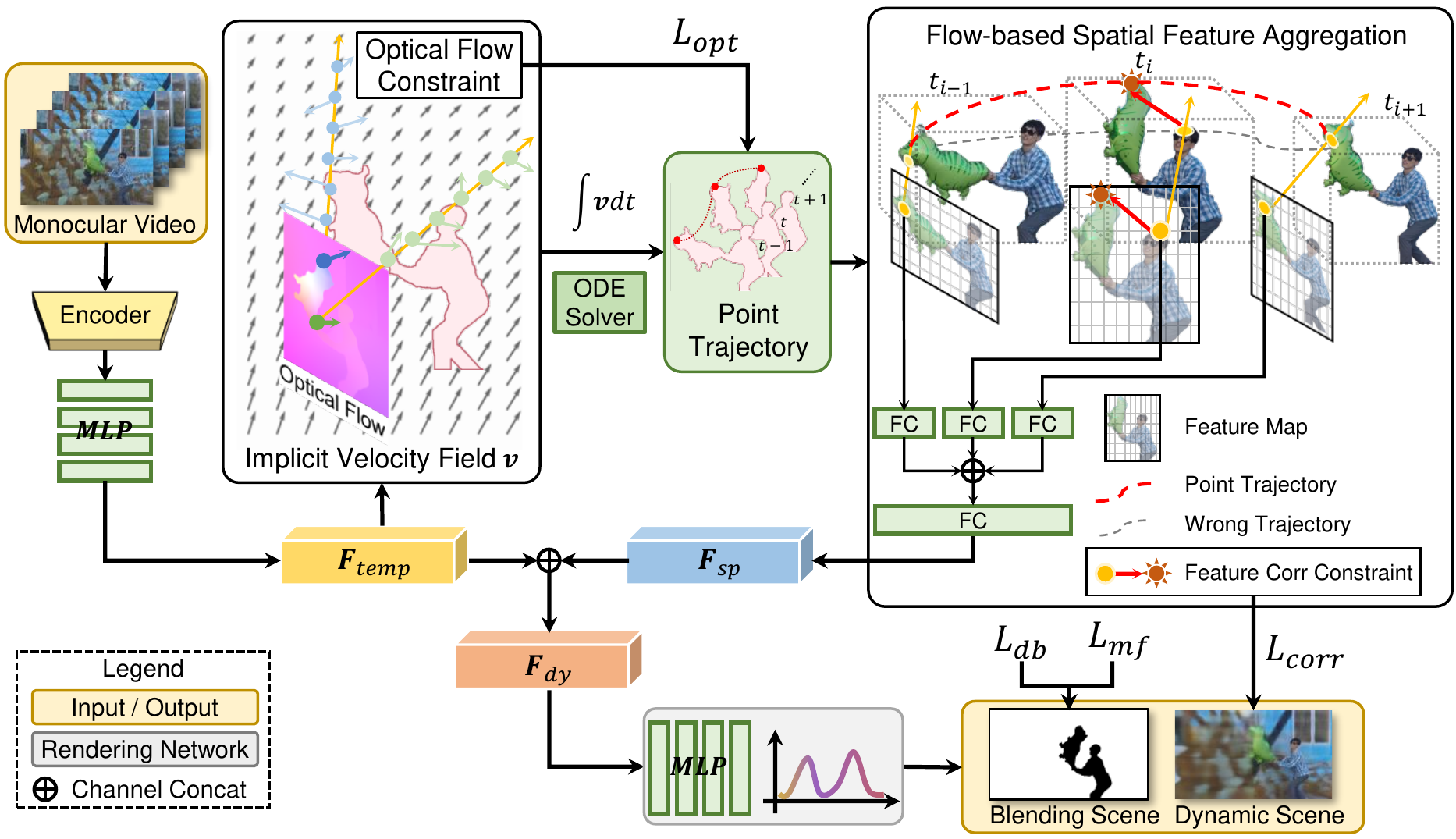}

   \caption{
   The overview of our generalizable dynamic field.
   We first exploit an implicit velocity field from the extracted temporal feature $\boldsymbol{F}_{temp}$. Then, we calculate point trajectory on the velocity field, and exploit the spatial feature $\boldsymbol{F}_{sp}$ with the proposed flow-based spatial feature aggregation module. We incorporate $\boldsymbol{F}_{temp}$ and $\boldsymbol{F}_{sp}$ as the point feature $\boldsymbol{F}_{dy}$ for rendering dynamic scene and design $L_{opt}$ and $L_{corr}$ to jointly optimize point features and trajectories. 
   }
   \label{fig:overview}
\end{figure*}
}

\section{Related Work}
\textbf{Novel view synthesis for static scenes.}
The long standing problem of novel view synthesis aims to construct new views of a scene
from multiple posed images. Early works needed dense views captured from the scene \cite{96siggraph/levoy_lightfr,96siggraph/steven_lumigraph}. Recent studies have 
shown great progress by explicitly representing 3D scenes as neural representations 
\cite{21cvpr/ricardo_nerfw,20eccv/ben_nerf,22cvpr/kangle_dsnerf,22cvpr/chen_hallucinated,22eccv/xiangli_bungeenerf,22cvpr/liu_neuralray,21iccv/jang_codenerf,21cvpr/srinivasan_nerv}. 
However, these methods train a separate model for each scene and need various training time for optimization. PixelNeRF \cite{21cvpr/yu_pixelnerf} and MVSNeRF \cite{21iccv/chen_mvsnerf} proposed feature-based methods that 
directly render new scenes from the encoded features. Additionally, many researchers 
studied the generalization and decomposition abilities of novel view synthesis model \cite{21cvpr/chen_pigan,20nips/schwarz_graf,21cvpr/niemeyer_giraffe,21iclr/gu/stylenerf,22cvpr/cai_pix2nerf,21iccv/jang_codenerf,21iccv/trevithick_grf}. Compared with these methods, our method studies the synthesis and generalization ability of dynamic scenes.

\textbf{Space-time view synthesis from monocular videos.}
With the development of neural radiance field in static 
scenes, recent studies started to 
address the dynamic scene reconstruction problem \cite{21cvpr/li_nsff,21cvpr/xian_stnif,21iccv/du_nerflow,22cvpr/li_neural3dvideo,21cvpr/ost_neuralscenegraph,20cvpr/dnerf}. In monocular videos, a key challenge is that there exist multiple scene constructions implying the same observed image sequences.
Previous approaches addressed this challenge by using 3D coordinates with time as point features and adding scene priors or spatio-temporal regularization. Scene priors such as shadow modeling \cite{22nips/wu_d2nerf}, human shape prior \cite{22cvpr/weng_humannerf,21siggraph_asia/liu_neuralactor}, and facial appearance \cite{21cvpr/guy_nerface} are
object-specific and closer to special contents. Spatio-temporal regularization such as depth and flow regularization \cite{21iccv/du_nerflow,21iccv/chen_dynerf,21cvpr/li_nsff,21cvpr/xian_stnif} and object deformations \cite{21iccv/tretschk_nr-nerf} are more object-agnostic, but weaker in applying consistency restriction to dynamic foregrounds. 
In this paper, we study the challenge by constraining point features.

\textbf{Scene flow for reconstruction.}
Scene flow firstly proposed by \cite{99iccv/vedula_sceneflow} depicts the motion of each point in the dynamic scene.
Its 2D projection \ie optical flow contributes to many downstream applications such as video super-resolution \cite{20tip/wang_ofrnet}, video tracking \cite{06joe/emanuele_videotrack}, video segmentation \cite{16cvpr/tsai_videoseg}, and video recognition \cite{17cvpr/quo_kinetics}. 
Several methods studied the scene flow estimation problem based on point cloud sequences \cite{iccv/niemeyer_oflow,19cvpr/liu_flownet3d}. However, estimating scene flow from an observed video is an ill-posed problem.
In dynamic scene reconstruction, previous works \cite{21cvpr/li_nsff,21iccv/chen_dynerf} established discrete scene flow for pixel consistency in observation time. In this way, the flow consistency in observed frames can only be constrained, 
leading to ambiguity in non-observed time. 
Du~\etal~\cite{21iccv/du_nerflow}~proposed to build the continuous flow field in a dynamic scene. Compared to these methods,
we study the generalization ability of the flow field in different scenes.

\section{Approach}
In this section, we introduce the proposed model that we term as \ourM. We first present the overview of our model and then detail the
approach of each part. Lastly, we introduce several strategies for optimizing the model.
\subsection{\ourM}
\label{sec:overview}
\ourM~aims to model generalizable radiance field of dynamic scenes.
Our method takes multiple monocular videos as input and uses optical flows, depth maps, and binary masks of foreground objects as supervision signals. 
Those signals could be generated by pretrained models automatically.
We build the generalizable static and dynamic fields for backgrounds and foregrounds separately.
For generalizable dynamic field, we suppose that spatio-temporal points flow along their trajectories and hold consistent features over time.
We first extract video features by a CNN encoder and generate the temporal feature vectors based on the extracted features.
Then, we build an implicit velocity field from the temporal feature vectors to estimate point trajectories.
Next, we exploit the estimated point trajectories as indexes to find the local image patches on each frame and 
extract the spatial features from the patches with the proposed flow-based feature aggregation module.
Different from NeRF \cite{20eccv/ben_nerf} that uses scene-agnostic point positions to render image color, we aggregate the scene-dependent spatial and
temporal features as the final point features to render foreground scenes with volume rendering. Finally, we design the optical flow and feature correspondence constraints for jointly learning point features and trajectories.
For generalizable static field, we suppose backgrounds are static and each video frame is considered as a different view of the background.
Since some background parts may be occluded by the foreground, we
design an effective strategy to sample background point features.
In the end, we combine the generalizable static and dynamic fields for rendering free-point videos.

\subsection{Generalizable Dynamic Field}
In this section, we introduce our generalizable dynamic field that renders novel views of dynamic foregrounds. We denote a monocular video as $\boldsymbol{V}=\{\boldsymbol{V}^i,i=1,2...,K\}$ consisting of $K$ frames . For each frame $\boldsymbol{V}^i$, $t_i$ is the observed timestamp and $\Pi^i$ is the projection transform from the world coordinate to the 2D frame pixel coordinate.
As \figref{fig:overview} shows, given a video $\boldsymbol{V}$, we exploit a video encoder $E_{dy}$ to extract the video feature vector represented as $E_{dy}(\boldsymbol{V})$ and generate the temporal feature vector
$\boldsymbol{F}_{temp}$ with a multiple layer perceptron (MLP) $W_{temp}$: $E_{dy}(\boldsymbol{V})\rightarrow \boldsymbol{F}_{temp}$.
We build the implicit velocity field based on $\boldsymbol{F}_{temp}$.

\textbf{Implicit velocity field.}
We suppose points are moving in the scene and
represent a spatio-temporal point as $\boldsymbol{p}=(\boldsymbol{x}_{p}, t_{p})$ including the 3D point position $\boldsymbol{x}_{p}$ and time $t_{p}$. We define $\boldsymbol{\Phi}$ as the continuous point trajectory and $\boldsymbol{\Phi}(\boldsymbol{p}, t)$ denotes the position of point $\boldsymbol{p}$ at timestamp $t$. We further define the velocity field $\boldsymbol{v}$ which includes the 3D velocity of each point. The relationship between point velocity and trajectory is specified as follows,
\begin{equation}
       \frac{\partial \boldsymbol{\Phi}(\boldsymbol{p},t)}{\partial t}=\boldsymbol{v}\Big(\boldsymbol{\Phi}(\boldsymbol{p},t), t \Big),\ \ 
    \text{ s.t. } \boldsymbol{\Phi}(\boldsymbol{p},t_{p})=\boldsymbol{x}_{p}.
\end{equation}
$\boldsymbol{v}$ is conditioned on $\boldsymbol{F}_{temp}$ and implemented by an MLP $W_{vel}$: $(\boldsymbol{F}_{temp}(\boldsymbol{V}), \boldsymbol{\Phi}, t) \rightarrow \boldsymbol{v}$. 
We calculate the point trajectory at an observed timestamp $t_{i}$ with Neural ODE \cite{18nips/chen_neuralode},
\begin{equation}
\label{eq:calculate_traj}
    \boldsymbol{\Phi}(\boldsymbol{p},t_{i})= \boldsymbol{x}_{p} + \int_{t_{p}}^{t_{i}} \boldsymbol{v}\big(\boldsymbol{\Phi}(\boldsymbol{p},t), t\big)dt.
\end{equation}
We take $\boldsymbol{\Phi}(\boldsymbol{p},t_{i})$ as an index to query the spatial feature. 

\textbf{Flow-based spatial feature aggregation.}
We employ $\Pi^{i}$ to project $\boldsymbol{\Phi}(\boldsymbol{p},t_{i})$ on $\boldsymbol{V}^{i}$ and find the local image patch indexed by the projected position $\Pi^{i}(\boldsymbol{\Phi}(\boldsymbol{p},t_{i}))$. 
We extract the feature vector $\boldsymbol{F}_{sp}^{i}$ from the patch with the encoder $E_{dy}$ and a fully connected layer $fc_1$,
\begin{equation}
    \boldsymbol{F}_{sp}^{i}(\boldsymbol{V}; \boldsymbol{p}) = fc_1\Big(E_{dy}\big(\boldsymbol{V}^{i}; \Pi^{i}(\boldsymbol{\Phi}(\boldsymbol{p},t_{i}))\big)\Big).
\end{equation}
In practice, $E_{dy}\big(\boldsymbol{V}^{i}; \Pi^{i}(\boldsymbol{\Phi}(\boldsymbol{p},t_{i}))\big)$ is implemented by using $E_{dy}$ to extract the frame-wise feature map of $\boldsymbol{V}^{i}$ and sampling the feature vector at $\Pi^{i}(\boldsymbol{\Phi}(\boldsymbol{p},t_{i}))$ with bilinear interpolation. The spatial feature vector $\boldsymbol{F}_{sp}$ is then calculated by incorporating $\{\boldsymbol{F}_{sp}^1, \boldsymbol{F}_{sp}^2,..., \boldsymbol{F}_{sp}^K\}$ with a fully connected layer $fc_2$,
\begin{equation}
\label{eq:sp_feat}
    \boldsymbol{F}_{sp}(\boldsymbol{V}; \boldsymbol{p}) = fc_2\big(\boldsymbol{F}_{sp}^{1}(\boldsymbol{V}; \boldsymbol{p}), \boldsymbol{F}_{sp}^{2}(\boldsymbol{V}; \boldsymbol{p}),...,\boldsymbol{F}_{sp}^{K}(\boldsymbol{V}; \boldsymbol{p})\big).
\end{equation}
Finally, we incorporate $\boldsymbol{F}_{temp}$ and $\boldsymbol{F}_{sp}$ as the point feature vector $\boldsymbol{F}_{dy}$,
\begin{equation}
    \boldsymbol{F}_{dy}(\boldsymbol{V}; \boldsymbol{p})=concat\{\boldsymbol{F}_{temp}(\boldsymbol{V}), \boldsymbol{F}_{sp}(\boldsymbol{V}; \boldsymbol{p})\}.
\end{equation}
We build the generalizable dynamic field based on $\boldsymbol{F}_{dy}$.

\textbf{Dynamic foreground rendering.}
Taking a point $\boldsymbol{p}$ and its feature vector $\boldsymbol{F}_{dy}(\boldsymbol{V}; \boldsymbol{p})$ as input, our generalizable dynamic radiance field $W_{dy}$: $(\boldsymbol{p}, \boldsymbol{F}_{dy}(\boldsymbol{V}; \boldsymbol{p})) \rightarrow (\boldsymbol{c}_{dy}, \sigma_{dy},b) $ predicts the volume density $\sigma_{dy}$, color $\boldsymbol{c}_{dy}$ and blending weight $b$ of the point.
We follow \DynNeRF~\cite{21iccv/chen_dynerf} and utilize $b$ to judge
whether a point belongs to static background or dynamic foreground. We exploit volume rendering to approximate each pixel color of an image. Concretely, given a ray $\boldsymbol{r}(u)=\boldsymbol{o}+u\boldsymbol{d}$ starting from the camera center $\boldsymbol{o}$ along the direction $\boldsymbol{d}$ through a pixel, its color is integrated by 
\begin{equation}
\label{eq:dynamic_rendering}
    \boldsymbol{C}_{dy}(\boldsymbol{r}) = \int_{u_n}^{u_f} T_{dy}(u)\sigma_{dy}(u)\boldsymbol{c}_{dy}(u)du,
\end{equation}
where $u_n, u_f$ are the bounds of the volume rendering depth range and $T_{dy}(u)=exp(-\int_{u_n}^u\sigma_{dy}(\boldsymbol{r}(s))ds$ is the accumulated transparency.
We simplify $\boldsymbol{c}(u)=\boldsymbol{c}(\boldsymbol{r}(u), \boldsymbol{d})$ and  $\sigma(u)=\sigma(\boldsymbol{r}(u))$ here and in the following sections.
Next we present the optical flow and feature correspondence constraints that jointly supervise our model.

\textbf{Optical flow constraint.} We supervise $\boldsymbol{v}$ with the optical flow $\boldsymbol{f}^{gt}$. In practice, it can only approximate the backward flow $\boldsymbol{f}_{bw}^{gt}$ and forward flow $\boldsymbol{f}_{fw}^{gt}$ between two consecutive video frames \cite{20eccv/zachary_raft}. 
We hence estimate the point backward and forward trajectory variations during the period that the point passes between two frames and calculate optical flows by integrating the trajectory variations of each point along camera rays.
Formally, given a ray $\boldsymbol{r}(u)$ 
through a pixel on the frame $\boldsymbol{V}^i$ at time $t_i$, for each point on the ray \ie $\boldsymbol{p}(u)=(\boldsymbol{r}(u),t_{i})$ the trajectory variations $\Delta \boldsymbol{\Phi}_{bw}$ back to $t_{i-1}$ and $\Delta \boldsymbol{\Phi}_{fw}$ forward to $t_{i+1}$ are obtained by the following equation,
\begin{equation}
\label{eq:traj_variation}
    \Delta \boldsymbol{\Phi}_{\{bw,fw\}}(\boldsymbol{p}(u))=\int_{t_{i}}^{t_{\{i-1,i+1\}}} \boldsymbol{v}\Big(\boldsymbol{\Phi}(\boldsymbol{p}(u),t),t\Big)dt.
\end{equation}
We follow previous works \cite{21cvpr/li_nsff,21iccv/chen_dynerf} and exploit the 
volume rendering to integrate pseudo optical flows $\boldsymbol{f}_{bw}$ and $\boldsymbol{f}_{fw}$ by utilizing the estimated volume density $\sigma_{dy}$ of each point,
\begin{equation}
    \boldsymbol{f}_{\{bw,fw\}}(\boldsymbol{r})=\int_{u_n}^{u_f} T_{dy}(u)\sigma_{dy}(u)\Delta \boldsymbol{\Phi}_{\{bw,fw\}}^{i}(\boldsymbol{p}(u)) du,
\end{equation}
where $\Delta\boldsymbol{\Phi}_{\{bw,fw\}}^{i}=\Pi^i(\Delta\boldsymbol{\Phi}_{\{bw,fw\}})$ denotes we use $\Pi^i$ to project 3D trajectory variations on the image plane of $\boldsymbol{V}^i$.
We supervise the pseudo flows by the ground truth flows,
\begin{equation}
    L_{opt}=\sum_{\boldsymbol{r}}\Big(\boldsymbol{f}_{\{bw,fw\}}(\boldsymbol{r})-\boldsymbol{f}_{\{bw,fw\}}^{gt}(\boldsymbol{r})\Big).
\end{equation}
In this way, the relation of point trajectories along a ray is limited by the optical flow supervision.

\textbf{Feature correspondence constraint.} According to \secref{sec:overview}, 
a point moving along the trajectory holds the same feature and represents the consistent color and density.
For each ray $\boldsymbol{r}_{curr}$ at the current time $t_i$ through a pixel of $\boldsymbol{V}^i$, we warp the ray from the neighboring
observed timestamps $t_{i-1}$ and $t_{i+1}$ as $\boldsymbol{r}_{bw}$ and $\boldsymbol{r}_{fw}$ separately by using \eqref{eq:traj_variation},
\begin{equation}
    \boldsymbol{r}_{\{bw,fw\}}(u) = \boldsymbol{r}_{curr}(u) + \Delta \boldsymbol{\Phi}_{\{bw,fw\}}(\boldsymbol{p}_{curr}(u)),
\end{equation}
where $\boldsymbol{p}_{curr}(u)=(\boldsymbol{r}_{curr}(u),t_{i})$.
We render the pixel color from the point features not only along the ray $\boldsymbol{r}_{curr}$ at the time $t_i$, but also along the wrapped rays  $\boldsymbol{r}_{bw}$ at $t_{i-1}$ and $\boldsymbol{r}_{fw}$ at $t_{i+1}$.
The predicted pixel color $\boldsymbol{C}_{dy}$ is rendered by using \eqref{eq:dynamic_rendering} and supervised by the ground truth colors $\boldsymbol{C}_{dy}^{gt}$,
\begin{equation}
    L_{\{bw,curr,fw\}}=\sum_{\boldsymbol{r}}||\boldsymbol{C}_{dy}(\boldsymbol{r}_{\{bw,curr,fw\}})-\boldsymbol{C}_{dy}^{gt}(\boldsymbol{r})||_2.
\end{equation}
The feature correspondence constraint $L_{corr}$ is defined as
\begin{equation}
    L_{corr} = L_{bw} + L_{curr} + L_{fw}.
\end{equation}
$L_{corr}$ supervises the predicted color and point features $\boldsymbol{F}_{dy}$. 

\subsection{Generalizable Static Field}
As mentioned in \secref{sec:overview}, for some background parts occluded by the changing foreground in the current view, their features implying the foreground cannot infer the correct background information. However, the occluded parts could be seen in non-occluded views with correct background features. To this end, given a point position $\boldsymbol{x}$, the background point feature vector $\boldsymbol{F}_{st}$ is produced by
\begin{equation}
    \boldsymbol{F}_{st}(\boldsymbol{V}; \boldsymbol{x}) = fc_3\Big(E_{st}(\boldsymbol{V}^*; \Pi^*(\boldsymbol{x}))\Big),
\end{equation}
where $fc_3$ is a fully connected layer and $E_{st}$ denotes the image encoder. $\boldsymbol{V}^*$ and $\Pi^*$ are the non-occluded frame and corresponding projection transform respectively.
$E_{st}(\boldsymbol{V}^*; \Pi^*(\boldsymbol{x}))$ denotes that we find the local image patch at $\Pi^*(\boldsymbol{x})$ and extract the feature vectors with an image encoder $E_{st}$ similar to $E_{dy}\big(\boldsymbol{V}^{i}; \Pi^{i}(\boldsymbol{\Phi}(\boldsymbol{p},t_{i}))\big)$.
Since there is no prior in which frames they can be
exposed, we apply a straightforward yet effective strategy by randomly sampling one frame from the other frames in the
video. We represent the static scene as a radiance field to infer the color $\boldsymbol{c}_{st}$ and density $\sigma_{st}$ by using an MLP $W_{st}$: $(\boldsymbol{x}, \boldsymbol{d}, \boldsymbol{F}_{st}(\boldsymbol{V}; \boldsymbol{x})) \rightarrow (\boldsymbol{c}_{st}, \sigma_{st})$.
The expected background color $\boldsymbol{C}_{st}$ is given by
\begin{equation}
    \boldsymbol{C}_{st}(\boldsymbol{r}) = \int_{u_n}^{u_f}T_{st}(u)\sigma_{st}(u)\boldsymbol{c}_{st}(u)du,
\end{equation}
where $T_{st}(t) = exp(-\int_{u_n}^u \sigma_{st}(\boldsymbol{r}(s)) ds$.
We employ foreground masks $M$ to optimize the static field by supervising the pixel color in each video frame in the background regions (where $M(\boldsymbol{r})=0$),
\begin{equation}
    L_{st} = \sum_{\boldsymbol{r}}||(\boldsymbol{C}_{st}(\boldsymbol{r}) - \boldsymbol{C}_{st}^{gt}(\boldsymbol{r}))(1-M(\boldsymbol{r}))||_2,
\end{equation}
where $\boldsymbol{C}_{st}^{gt}$ represents the ground truth color.
\CheckRmv{
\begin{table*}[!hbt]
\centering
\caption{Novel view synthesis on training frames from single videos. While this setting does not require cross-scene generalization ability, our \ourM~still achieves better performance.}
\label{tab:results_single_video}
\resizebox{\textwidth}{!}{%
\begin{tabular}{lcccccccc}
\hline
PSNR $\uparrow$ / LPIPS $\downarrow$     & Jumping       & Skating       & Truck         & Umbrella      & Balloon1      & Balloon2      & Playground    & Average       \\ \hline
NeRF \cite{20eccv/ben_nerf}           & 20.58 / 0.305 & 23.05 / 0.316 & 22.61 / 0.225 & 21.08 / 0.441 & 19.07 / 0.214 & 24.08 / 0.098 & 20.86 / 0.164 & 21.62 / 0.252 \\
NeRF \cite{20eccv/ben_nerf} + time    & 16.72 / 0.489 & 19.23 / 0.542 & 17.17 / 0.403 & 17.17 / 0.752 & 17.33 / 0.304 & 19.67 / 0.236 & 13.80 / 0.444 & 17.30 / 0.453 \\
Yoon \etal~\cite{20cvpr/yoon_dsvdmv} & 20.16 / 0.148 & 21.75 / 0.135 & 23.93 / 0.109 & 20.35 / 0.179 & 18.76 / 0.178 & 19.89 / 0.138 & 15.09 / 0.183 & 19.99 / 0.153 \\
Tretschk~\etal~\cite{21iccv/tretschk_nr-nerf} & 19.38 / 0.295 & 23.29 / 0.234 & 19.02 / 0.453 & 19.26 / 0.427 & 16.98 / 0.353 & 22.23 / 0.212 & 14.24 / 0.336 & 19.20 / 0.330 \\
Li~\etal~\cite{21cvpr/li_nsff}       & 24.12 / 0.156 & 28.91 / 0.135 & 25.94 / 0.171 & 22.58 / 0.302 & 21.40 / 0.225 & 24.09 / 0.228 & 20.91 / 0.220 & 23.99 / 0.205 \\
NeuPhysics~\cite{22nips/qiao_neuphysics}    & 20.16 / 0.205 & 25.13 / 0.166 & 22.62 / 0.212 & 21.02 / 0.426 & 16.68 / 0.238 & 22.54 / 0.265 & 15.10 / 0.367 & 20.48 / 0.282 \\ 
\DynNeRF~\cite{21iccv/chen_dynerf}       & 24.23 / 0.144 & 28.90 / 0.124 & 25.78 / 0.134 & 23.15 / \textbf{0.146} & 21.47 / \textbf{0.125} & 25.97 / 0.059 & \textbf{23.65} / \textbf{0.093} & 24.74 / 0.118 \\ \hline
\ourM                              & \textbf{24.26} / \textbf{0.091} & \textbf{32.06} / \textbf{0.044} & \textbf{27.56} / \textbf{0.115} & \textbf{23.62} / 0.180 & \textbf{21.89} / 0.129 & \textbf{27.36} / \textbf{0.052} & 22.61 / 0.130 & \textbf{25.62} / \textbf{0.106} \\ \hline
\end{tabular}%
}
\end{table*}
}
\CheckRmv{
\begin{figure*}[t]
  \centering
   \includegraphics[width=0.97\linewidth]{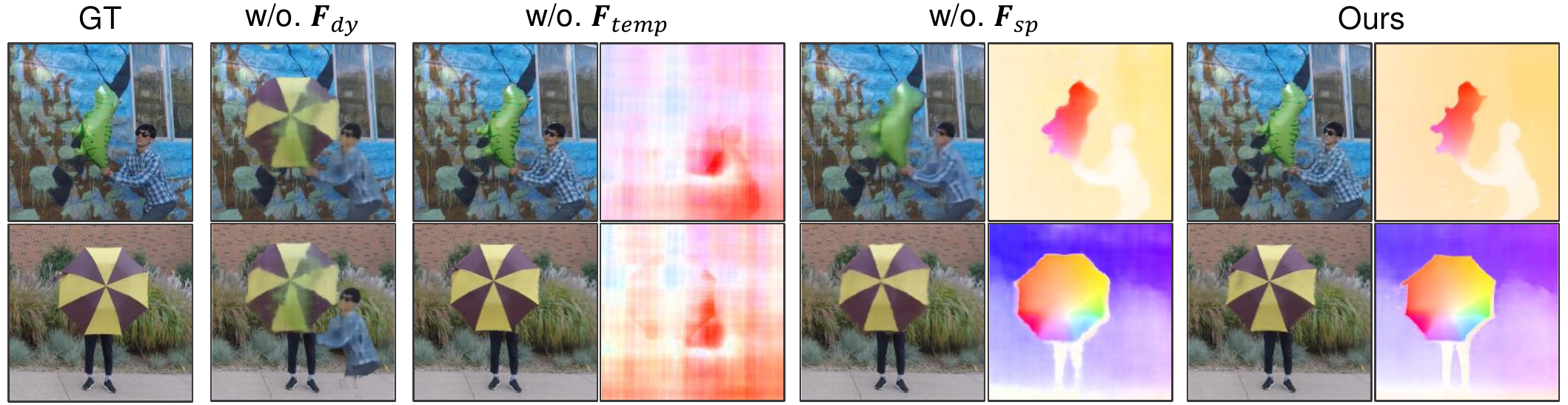}

   \caption{Qualitative results of jointly optimizing Balloon2 and Umbrella scenes. The foregrounds of two scenes are mixed without $\boldsymbol{F}_{dy}$. Our method renders more accurate novel views and predicts plausible scene flows (listed beside the RGB images) by incorporating $\boldsymbol{F}_{temp}$ and $\boldsymbol{F}_{sp}$.}
   \label{fig:balloon2_umbrella}
\end{figure*}
}
\subsection{Optimization}

The final dynamic scene color combines the colors from the generalizable dynamic and static fields,
\begin{equation}
\begin{split}
    \boldsymbol{C}_{full}(\boldsymbol{r})=&\int_{u_n}^{u_f}T_{full}(u)\sigma_{full}(u)\boldsymbol{c}_{full}(u) du,
\end{split}
\end{equation}
where
\begin{equation}
    \sigma_{full}(u)\boldsymbol{c}_{full}(u)=(1-b)\sigma_{st}(u)\boldsymbol{c}_{st}(u) +b\sigma_{dy}(u)\boldsymbol{c}_{dy}(u),
\end{equation}
and applies the reconstruction loss,
\begin{equation}
    L_{full}=\sum_{\boldsymbol{r}}||\boldsymbol{C}_{full}(\boldsymbol{r})-\boldsymbol{C}_{full}^{gt}(\boldsymbol{r})||_2.
\end{equation}
Next we design several strategies to optimize our model.

\textbf{Point trajectory discretization.}
While point trajectory can be numerically estimated by the continuous integral \eqref{eq:calculate_traj} with Neural ODE solvers \cite{18nips/chen_neuralode}, it needs plenty of time for querying each point trajectory. To accelerate the process, we propose to partition $[t_{p}, t_{i}]$ into $N$ evenly-spaced bins
and suppose the point velocity in each bin is a constant. The point trajectory can be estimated by the following equation,
\begin{equation}
\label{eq:discrete_traj}
    \boldsymbol{\Phi}(\boldsymbol{p},t_{i})= \boldsymbol{x}_{p} + \sum_{n=1}^{N} \boldsymbol{v}\big(\boldsymbol{\Phi}(\boldsymbol{p},t+\Delta t), t+\Delta t\big)\Delta t,
\end{equation}
where $\Delta t=\frac{n}{N}(t_{i}-t_p)$. 
\CheckRmv{
\begin{figure}[!]
  \centering
   \includegraphics[width=0.95\linewidth]{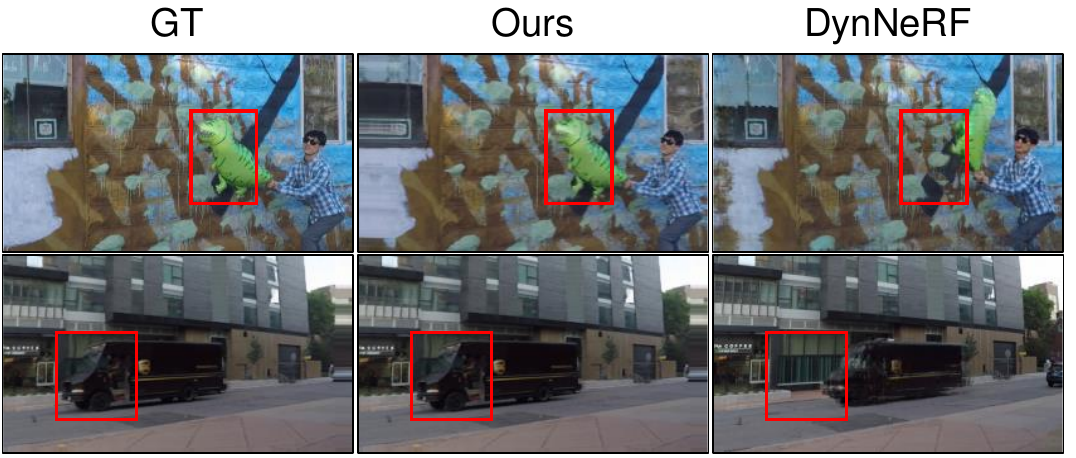}

   \caption{Novel view synthesis on unseen frames. Compared to the ground truths, our model could transfer to new motions, whereas \DynNeRF~\cite{21iccv/chen_dynerf} only interpolates in the training frames.}
   \label{fig:unseen_frames}
\end{figure}
}
\CheckRmv{
\begin{figure*}[t]
  \centering
   \includegraphics[width=0.97\linewidth]{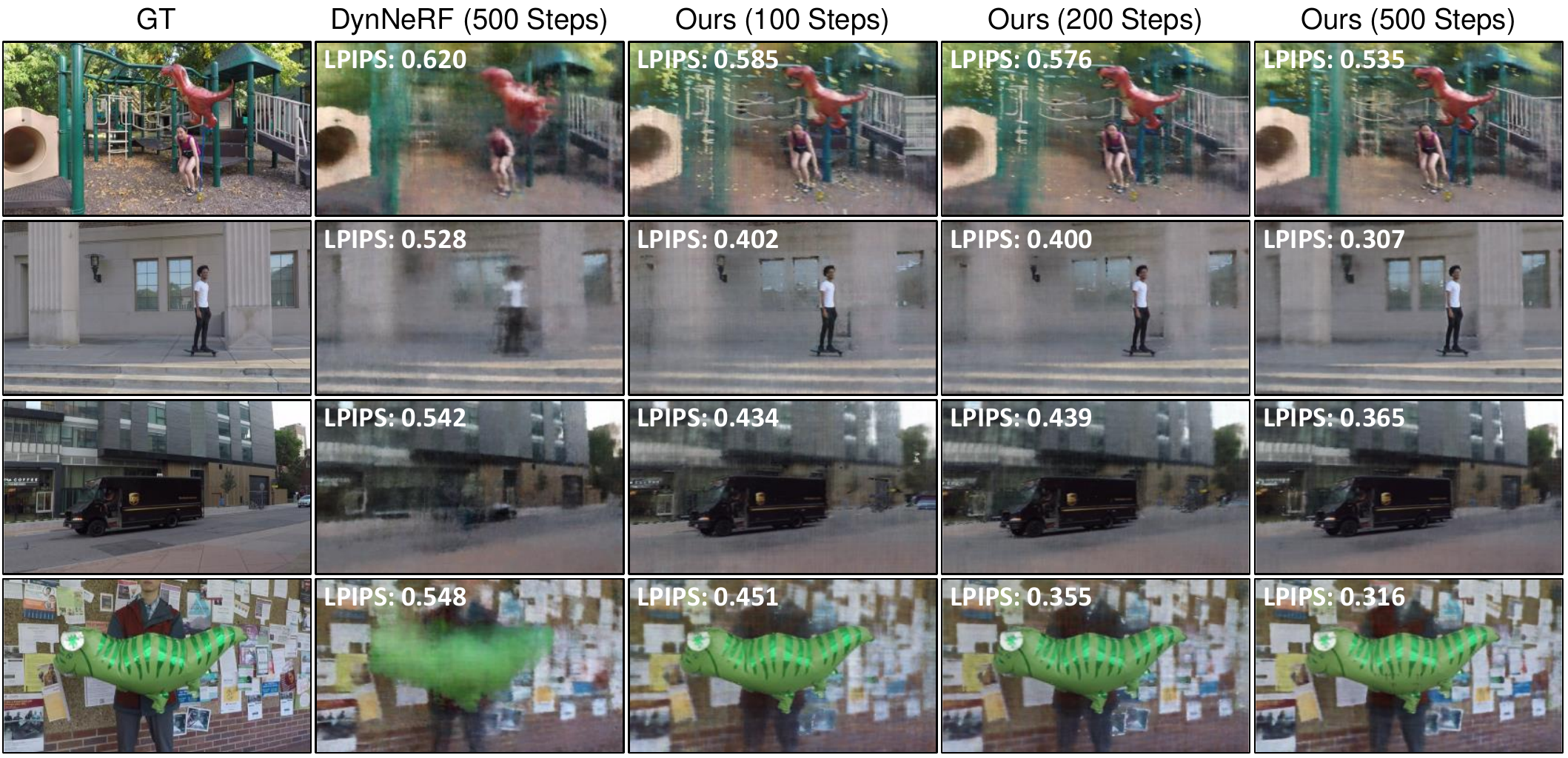}

   \caption{Fast novel scene adaption. We used the pretrained model on Balloon2 scene to fine-tune the novel views of unseen dynamic scenes. We employed LPIPS \cite{18cvpr/zhang_lpips} (the lower is better) to evaluate the image similarity, which provides more correlation with human judgment than other indexes.
   }
   \label{fig:unseen_videos}
\end{figure*}
}

\CheckRmv{
\begin{table}[]
\centering
\caption{Quantitative results of novel view synthesis on unseen frames. We used the first four frames for training and tested the performance on the rest eight frames.}
\label{tab:unseen_frames}
\resizebox{\linewidth}{!}{%
\begin{tabular}{lccc}
\hline
Balloon2   / Truck                 & PSNR $\uparrow$         & SSIM $\uparrow$        & LPIPS $\downarrow$      \\ \hline
NeRF \cite{20eccv/ben_nerf}           & 20.33 / 20.26         & 0.662 / 0.669           & 0.224 / 0.256 \\
NeRF \cite{20eccv/ben_nerf} + time    & 20.22 / 20.26          & 0.661 / 0.639          & 0.218 / 0.256 \\
NeuPhysics~\cite{22nips/qiao_neuphysics}    & 19.45 / 20.24          & 0.478 / 0.517          & 0.343 / 0.285 \\
\DynNeRF~\cite{21iccv/chen_dynerf}    & 19.99 / 20.33          & 0.641 / 0.621          & 0.291 / 0.273 \\
MonoNeRF                               & \textbf{21.30 / 23.74} & \textbf{0.669 / 0.702} & \textbf{0.204 / 0.174} \\ \hline
\end{tabular}%
}
\end{table}
}
\textbf{Depth-blending restriction.} To calculate the blending weights within a foreground ray in the generalizable dynamic field, only the points in close proximity to the estimated ray depth are regarded as foreground points, whereas points beyond this range are excluded. We penalize the blending weights of
non-foreground points in foreground rays. Specifically, given a ray $\boldsymbol{r}(u)$ through a pixel and the pixel depth $u_d$,
We penalize the blending weights of the points on the ray out of the interval $(u_d-\epsilon, u_d+\epsilon)$,
\begin{equation}
    L_{db}=\sum_{u\in (u_n, u_d-\epsilon)\cup(u_d+\epsilon, u_f)}||b(\boldsymbol{r}(u))||_2,
\end{equation}
where $b(\boldsymbol{r}(u))$ is the blending weight at the position $\boldsymbol{r}(u)$. $\epsilon$ controls the surface thickness of the dynamic foreground.

\textbf{Mask flow loss.} We further constrain the consistency of  point features by minimizing blending weight variations along trajectories,
\begin{equation}
    L_{mf} = \sum_{i,j\in \{1,2,.., K\}}||b(\boldsymbol{\Phi}(\boldsymbol{p},t_i))-b(\boldsymbol{\Phi}(\boldsymbol{p},t_j))||_2,
\end{equation}
where $b(\boldsymbol{\Phi}(\boldsymbol{p},t))$ denotes the blending weight at $\boldsymbol{\Phi}(\boldsymbol{p},t)$.

\textbf{Other regularization.} We follow previous works \cite{21cvpr/li_nsff,21iccv/chen_dynerf,21iccv/du_nerflow} to use the depth constraint, sparsity constraint, and motion regularization and smoothness to train the model.

\section{Experiments}
In this section, we conducted experiments on the Dynamic Scene dataset \cite{20cvpr/yoon_dsvdmv}.
We first tested the performance of synthesizing novel views from single videos, and then we tested the generalization ability from multiple videos.
In the end, we carried out ablation studies on our model.
\subsection{Experimental Setup}
\textbf{Dataset.} We used the Dynamic Scene dataset \cite{20cvpr/yoon_dsvdmv} to 
evaluate the proposed method.  Concretely, it contains 9 video sequences that are captured with 12 cameras by using a camera rig. We followed previous works \cite{21cvpr/li_nsff, 21iccv/chen_dynerf} and derived each frame of the video from different cameras to simulate the camera motion. All the cameras capture images at 12 different timestamps $\{ t_i, i=1,2...,12\}$. Each training video contains twelve frames sampled from the $i^{th}$ camera at time $t_i$ followed \DynNeRF~\cite{21iccv/chen_dynerf}. 
We used COLMAP \cite{16cvpr/lutz_sfm,16eccv/lutz_mvs} to approximate the camera poses. It is assumed intrinsic parameters of all the cameras are the same. 
DynamicFace sequences were excluded because COLMAP fails to estimate camera poses.
All video frames were resized to 480$\times$270 resolution. We generated the depth, mask, and optical flow signals from the depth estimation model \cite{22pami/ranftl_depth}, Mask R-CNN \cite{17iccv/he_maskrcnn}, and RAFT \cite{20eccv/zachary_raft}.
\CheckRmv{
\begin{table}[]
\centering
\caption{Ablation studies on $\boldsymbol{F}_{dy}$, $\boldsymbol{F}_{temp}$, and $\boldsymbol{F}_{sp}$ by jointly optimizing Balloon2 and Umbrella scenes.}
\label{tab:abs_sp_temp}
\resizebox{\linewidth}{!}{%
\begin{tabular}{c|ccc}
\hline
PSNR $\uparrow$ /   LPIPS $\downarrow$ & Umbrella      & Balloon2      & Average          \\ \hline
w/o. $\boldsymbol{F}_{dy}$    & 20.59 / 0.256 & 22.79 / 0.159 & 21.57 / 0.230 \\ 
w/o. $\boldsymbol{F}_{temp}$   & 22.75 / 0.175 & 24.85 / 0.098 & 23.80 / 0.137 \\
w/o. $\boldsymbol{F}_{sp}$     & 22.81 / 0.181 & 25.09 / 0.143 & 23.95 / 0.162 \\ \hline
Ours           & \textbf{23.44} / \textbf{0.169} & \textbf{25.44} / \textbf{0.093} & \textbf{24.44} / \textbf{0.131} \\ \hline 
\end{tabular}%
}
\end{table}
}

\textbf{Implementation details.}
We followed pixelNeRF \cite{21cvpr/yu_pixelnerf} and used ResNet-based MLPs as our implicit velocity field $W_{vel}$ and rendering networks $W_{dy}$ and $W_{st}$. For generalizable dynamic field, we utilized Slowonly50 \cite{19iccv/he_slowfast} pretrained on Kinetics400 \cite{17cvpr/quo_kinetics} dataset as the encoder $E_{dy}$ with the frozen weights. We removed the final fully connected layer in the backbone and incorporated the first, second, and third feature layers for querying $\boldsymbol{F}_{sp}$. We simplified \eqref{eq:sp_feat} that $\boldsymbol{F}_{sp}$ at time $t_i$ is sampled from $\{\boldsymbol{V}^{i-1}, \boldsymbol{V}^i, \boldsymbol{V}^{i+1}\}$.
For generalizable static field, we used ResNet18 \cite{16cvpr/he_resnet} pretrained on ImageNet \cite{09cvpr/deng_imagenet} as the encoder $E_{st}$. We extracted a feature pyramid from video frames for querying $\boldsymbol{F}_{st}$. The vector sizes of $\boldsymbol{F}_{temp}$, $\boldsymbol{F}_{sp}$, $\boldsymbol{F}_{dy}$, and 
$\boldsymbol{F}_{st}$ are 256. Please refer to the \supp~for more details.
\CheckRmv{
\begin{figure}[t]
  \centering
   \includegraphics[width=0.95\linewidth]{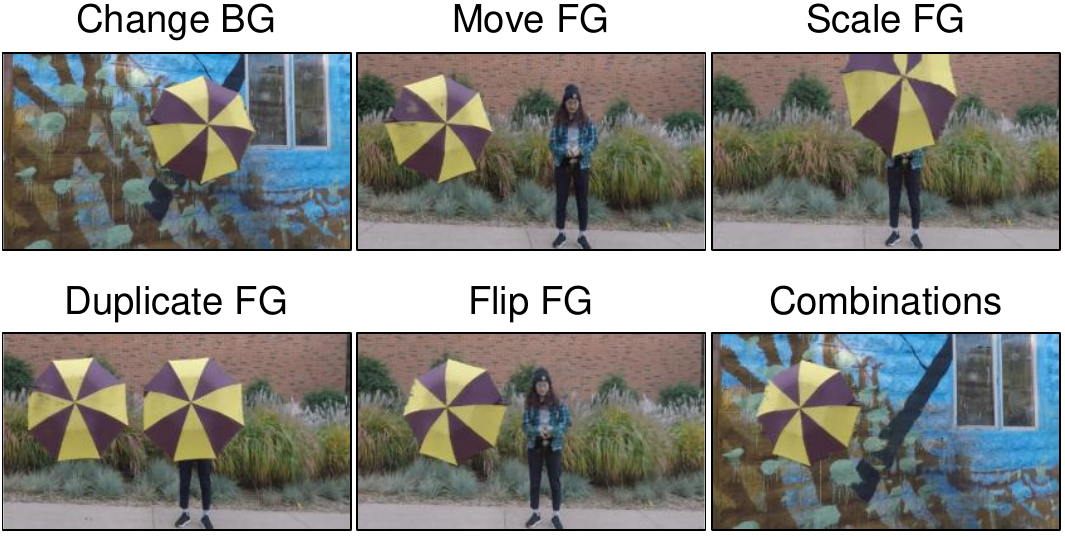}

   \caption{Our model supports many
   scene editing applications such as background changing, foreground moving, scaling, duplicating, flipping, and arbitrary combinations.
   }
   \label{fig:scene_editing}
\end{figure}
}
\begin{table}[t]
\centering
\caption{Numeric comparisons on $L_{db}, L_{mf}, \boldsymbol{F}_{st}$, and random sampling strategy.}
\label{tab:abs_db_mf_random_static}
\resizebox{\linewidth}{!}{%
\begin{tabular}{c|c|c|c}
\hline
PSNR $\uparrow$ / LPIPS $\downarrow$  & Training frames         & Unseen frames     & Unseen videos \\\hline
w/o. $L_{db}$                & 22.76 / 0.148  & 20.69 / 0.348  & 21.68 / 0.345
         \\\hline
w/o. $L_{mf}$                & 22.95 / 0.135   & 20.95 / 0.354 & 21.67 / 0.342    \\\hline
w/o. $\boldsymbol{F}_{st}$ &  21.75 / 0.230  &  19.79 / 0.371 &  21.81 / 0.337  \\\hline
w/o. \emph{random}        & 17.30 / 0.435       & 16.93 / 0.512 &  20.28 / 0.353    \\\hline     
Ours         & \textbf{23.02 / 0.130}   & \textbf{21.30 / 0.304} &    \textbf{22.63 / 0.277} \\ \hline 
\end{tabular}%
}
\end{table}
\CheckRmv{
\begin{table}[t]
\centering
\caption{Comparisons of solving the continuous trajectory by using ODE solver \cite{18nips/chen_neuralode} and our discretization method.}
\label{tab:ablation_ode}
\resizebox{\linewidth}{!}{%
\begin{tabular}{c|c|c|c}
\hline
Method       & $N=1$           & $N=2$              & ODE solver \cite{18nips/chen_neuralode}            \\ \hline
PSNR $\uparrow$ / LPIPS $\downarrow$  & 22.90 / 0.136 & 22.97 / 0.134  & \textbf{23.75 / 0.129} \\\hline
\end{tabular}%
}
\end{table}
}
\subsection{Novel View Synthesis from Single Video}
In this section, we trained the models from single monocular videos, where existing methods are applicable to this setting. Specifically, we first tested the performance on training frames, which is the widely-used setting to evaluate video-based NeRFs.
Then, we tested the generalization ability on unseen frames 
in the video, where other existing methods are not able to transfer well to the unseen motions.

\textbf{Novel view synthesis on training frames.}
To evaluate the synthesized novel views, we followed \DynNeRF~\cite{21iccv/chen_dynerf} and tested the performance by fixing the view to the first camera and changing time. 
We reported the PSNR and LPIPS \cite{18cvpr/zhang_lpips} in \tabref{tab:results_single_video}. We evaluated the performance of Li \etal~\cite{21cvpr/li_nsff}, Tretschk \etal~\cite{21iccv/tretschk_nr-nerf}, NeuPhysics~\cite{22nips/qiao_neuphysics}, and Chen \etal~\cite{21iccv/chen_dynerf} from the official implementations. Even without the need of generalization ability, our method achieves better results.

\textbf{Novel view synthesis on unseen frames.}
We split the video frames into two groups: four front frames were used for training and the rest of eight unseen frames were utilized to render novel views. \figref{fig:unseen_frames} shows that our model successfully renders new motions in the unseen frames, while \DynNeRF~\cite{21iccv/chen_dynerf} only interpolates novel views in the training frames. The reported PSNR, SSIM \cite{04tip/wang_ssim}, and LPIPS scores in \tabref{tab:unseen_frames} quantitatively verify the superiority of our model.
\subsection{Novel View Synthesis from Multiple Videos}
In the section, we tested the novel view synthesis performance on multiple dynamic scenes. It is worth noting that as existing methods can only learn from single monocular videos, they are not applicable to the settings that need to train on multiple videos. Then, we evaluated the novel scene adaption ability on several novel monocular videos.
Lastly, we conducted a series of scene editing experiments. 

\textbf{Novel view synthesis on training videos.}
We selected Balloon2 and Umbrella scenes to train our model. As shown in \figref{fig:balloon2_umbrella} and \tabref{tab:abs_sp_temp},
our model could distinguish foregrounds from two scenes and perform well with $\boldsymbol{F}_{dy}$.
Specifically, it predicts generalizable scene flows with $\boldsymbol{F}_{temp}$ and renders more accurate details with $\boldsymbol{F}_{sp}$.
\CheckRmv{
\begin{figure}[t]
  \centering
   \includegraphics[width=0.95\linewidth]{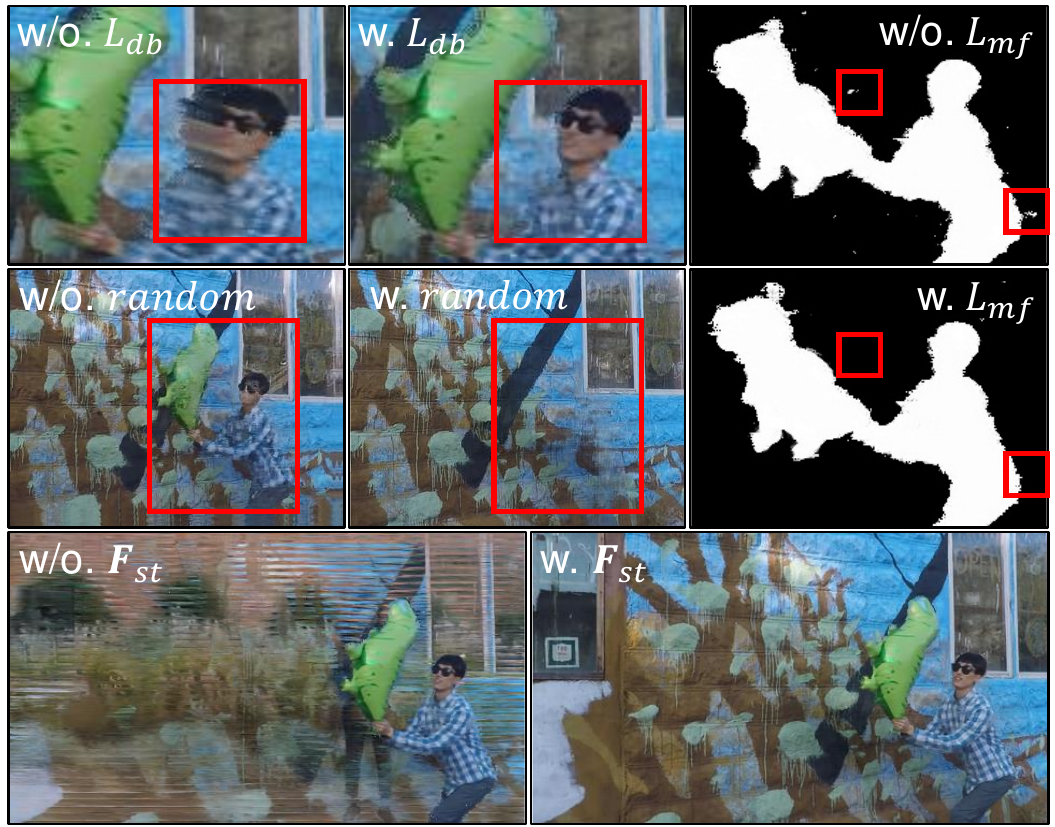}

   \caption{Ablation studies on $L_{db}$, $L_{mf}$, random sampling, and $\boldsymbol{F}_{st}$.}
   \label{fig:ablation_df_fl_loss}
\end{figure}
}

\textbf{Novel view synthesis on unseen videos.}
We explored the generalization ability of our model by pretraining on the Balloon2 scene and fine-tuning the pretrained model on other scenes. \figref{fig:unseen_videos} presents the results of four unseen videos: Playground, Skating, Truck, and Balloon1. We also pretrained \DynNeRF~\cite{21iccv/chen_dynerf} on 
the Balloon2 scene for a fair comparison. 
While \DynNeRF~only learns to render new scenes from scratch, our model transfers to scenes with correct dynamic motions. By further training with 500 steps, our model achieves better image rendering quality and higher LPIPS scores, which takes about 10 minutes.

\textbf{Scene editing.}
As \figref{fig:scene_editing} shows, our model further supports many scene editing applications by
directly processing point features without extra training. Changing background was implemented by exchanging the static scene features between two scenes. Moving, scaling, duplicating and flipping foreground were directly applied by operating video features in dynamic radiance field.
The above applications could be combined arbitrarily.
\subsection{Ablation Study}
We conducted a series of experiments to examine the proposed $L_{db}$, $L_{mf}$, $\boldsymbol{F}_{st}$, random sampling strategy, and trajectory discretization method.
We show in \figref{fig:ablation_df_fl_loss} that $L_{db}$ deblurs the margin of the synthesized foreground in novel views, $L_{mf}$ keeps the blending consistency in different views for delineating the foreground more accurately, The random sampling strategy successfully renders static scene without dynamic foreground, and backgrounds of two scenes are mixed without $\boldsymbol{F}_{st}$. We also show the numeric comparisons of $L_{db}, L_{mf}, \boldsymbol{F}_{st}$, and random sampling strategy in \tabref{tab:abs_db_mf_random_static}. 
In \tabref{tab:ablation_ode}, our discretization method reaches comparable results with ODE solvers \cite{18nips/chen_neuralode}
and achieves higher performance with the increase of $N$.
\section{Conclusion}
In this paper, we study the generalization ability of novel view synthesis from monocular videos. 
The challenge is the ambiguity of 3D point features and scene flows 
in the viewing directions. 
We find that video frame features and optical flows are a pair of complementary constraints for learning 3D point features and scene flows. To achieve this, we propose a generalizable dynamic radiance field called \ourM.
We estimate point features and trajectories from the video features extracted by a video encoder and render dynamic scenes from points features.
Experiments show that our model could learn a generalizable radiance field for dynamic scenes, and support many scene editing applications.

\textbf{Acknowledgements.}
This work was supported by the National Key Research and Development Program of China under Grant No. 2020AAA0108100, the National Natural Science Foundation of China under Grant No. 62206147 and 61971343. The authors would like to thank the Shanghai Artificial Intelligence Laboratory for its generous support of computational resources.
{\small
\bibliographystyle{ieee_fullname}
\bibliography{egbib}
}
\clearpage
\appendix
\begin{table}[]
\centering
\caption{Configurations for generalizable dynamic field.}
\label{tab:dynamic field params}
\resizebox{\linewidth}{!}{%
\begin{tabular}{l|cccccc}
parameter & $\alpha_{full}$  & $\alpha_{opt}$  & $\alpha_{corr}$ & $\alpha_{db}$  & $\alpha_{mf}$ & $\epsilon$  \\ \hline
value     & 1    & 0.02     & 4   & 0.01     & 1              & 0.03 
\end{tabular}%
}
\end{table}

\section{Implementation Details}
In this section, we present specific implementation details of our model and each experimental setting.
The entire model was trained on a NVIDIA A100 GPU with a total batch size of 1024 rays. The learning rate is 0.0005 without decaying. The initial training time is about 3 hours.

\subsection{Generalizable Dynamic Field}
We trained the generalizable dynamic field in an end-to-end manner. The overall loss function is
\begin{equation}
    L=\alpha_{full} L_{full} + \alpha_{opt} L_{opt} + \alpha_{corr} L_{corr} + \alpha_{db} L_{db} + \alpha_{mf} L_{mf}.
\end{equation}
The hyper-parameter values of all loss functions
are listed in \tabref{tab:dynamic field params}.
$\epsilon$ is the blending thickness as described in the $L_{db}$ in the paper.
We used the Slowonly50 network in SlowFast \cite{19iccv/he_slowfast} as the video encoder $E_{dy}$, which was pretrained on the Kinetics400 \cite{17cvpr/quo_kinetics} dataset, and removed all the temporal pooling layers in the network. We froze the weights of the pretrained model.
The temporal features $\boldsymbol{F}_{temp}(\boldsymbol{V})$ are the latent vectors of size 256 extracted by the encoder $E_{dy}$ and fused by $W_{dy}$, and the spatial features $\boldsymbol{F}_{st}(\boldsymbol{V}; \boldsymbol{p})$ of each point were extracted prior to the first 3 spatial pooling layers, which were upsampled using bilinear interpolation, concatenated in the channel dimension and fused with the fully connected layers to form latent vectors of size 256.
To incorporate the point feature into NeRF \cite{20eccv/ben_nerf} network, we followed pixelNeRF \cite{21cvpr/yu_pixelnerf} to use 
multi-layer perceptron (MLP) with residual modulation \cite{16cvpr/he_resnet} as our basic block. We employed 4 residual blocks to implement our implicit velocity field, and another 4 residual blocks as the rendering network.
\begin{figure}[]
  \centering
   \includegraphics[width=0.95\linewidth]{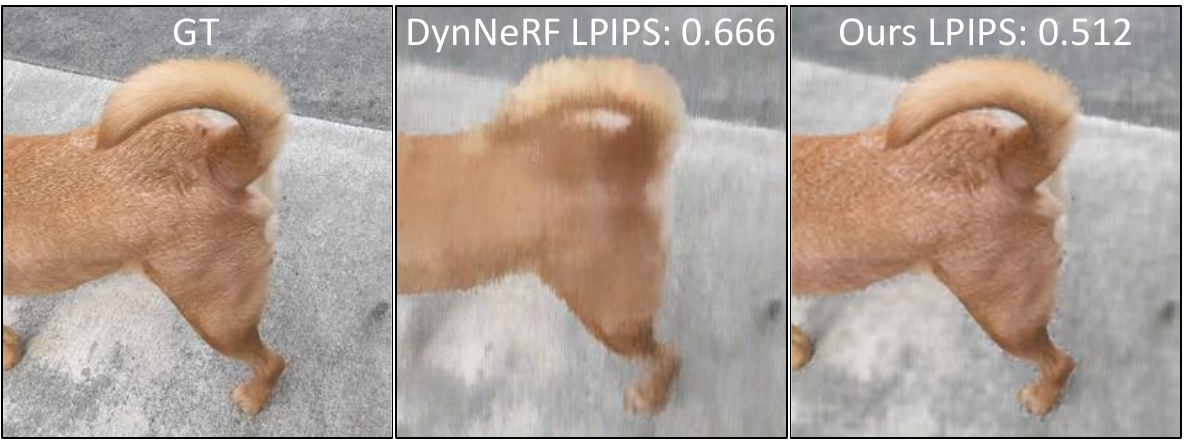}

   \caption{Qualitative results on the Nerfies~\cite{21iccv/park_nerfies} dataset .}
   \label{fig:nerfies}
\end{figure}
\begin{table}[]
\centering
\caption{Quantitative results on the Nerfies \cite{21iccv/park_nerfies} dataset.}
\label{tab:nerfies}
\resizebox{0.95\linewidth}{!}{%
\begin{tabular}{c|c|c|c}
\hline
 & NeRF [28] + time       & DynNeRF [13]   & MonoNeRF  \\ \hline
PSNR $\uparrow$ / LPIPS $\downarrow$   & 23.80  / 0.684 & 25.80  / 0.671 &  \textbf{27.77  / 0.501} \\\hline   

\end{tabular}%
}
\end{table}

\subsection{Generalizable Static Field}
We used ResNet18 \cite{16cvpr/he_resnet} as the image encoder $E_{st}$, which was pretrained on ImageNet \cite{09cvpr/deng_imagenet}. The point features $\boldsymbol{F}_{st}$ in generalizable static field were incorporated prior to 4 pooling layers in ResNet18, which were upsampled and concatenated to form latent vectors of size 256 aligned to each point. We also used MLP with residual modulation as basic architecture for static rendering network. To train the generalizable static field, we used the segmentation mask preprocessed by DynNeRF \cite{21iccv/chen_dynerf} and optimized the static field by using the image pixels that belong to static background.

\subsection{Novel View Synthesis on Unseen Frames}
Novel view synthesis on unseen frames aims to test the generalizable ability of our model on unseen motions in a fixed static scene. We used the first 4 frames to train our generalizable dynamic field, and evaluated the performance on the rest 8 frames for each scene.
The training step was set to 40000 in this setting.

\subsection{Novel View Synthesis on Unseen Videos}
Novel view synthesis on unseen videos aims to test the generalizable ability of our model on novel dynamic scenes. We pretrained our model on Balloon2 scene and finetuned the model on other scenes with the pretrained parameters. The pretraining step on Balloon2 scene was set to 20000. We used the official model \cite{21iccv/chen_dynerf} trained on Balloon2 scene for a fair comparison. The initial training time is about 3 hours for one scene, and the finetuning time is about 10 minutes.

\subsection{Scene Editing}
All the scene editing operations are conducted directly on the extracted backbone features without extra training. \textbf{Changing background} was conducted by exchanging the extracted image features in static field in our model. \textbf{Moving foreground} was implemented by moving the video features in the dynamic field at the corresponding position. \textbf{Scaling foreground} was implemented by scaling the video features in the dynamic field. \textbf{Duplicating foreground} was conducted by copying the video features to the corresponding position. \textbf{Flipping foreground} was applied by flipping the video features. Since the above operations are independent to each other, they can be combined in an arbitrary way.

\section{Generalization Results across Datasets}
We pretrained our model on the Dynamic Scene dataset \cite{20cvpr/yoon_dsvdmv} and fine-tuned the model on the Nerfies~\cite{21iccv/park_nerfies} dataset  with 500 steps, which contains videos recorded by cellphone cameras. Both \tabref{tab:nerfies} and \figref{fig:nerfies} show that MonoNeRF presents stronger generalization ability on cellphone videos.  
\end{document}